\documentclass[10pt,twocolumn,letterpaper]{article}

\usepackage{cvpr}              % To produce the CAMERA-READY version
\usepackage[dvipsnames]{xcolor}

\definecolor{cvprblue}{rgb}{0.21,0.49,0.74}
\usepackage[pagebackref,breaklinks,colorlinks,citecolor=cvprblue]{hyperref}

\usepackage{xspace}
\usepackage{xcolor}   
\usepackage{amssymb}
\usepackage{pifont}
\usepackage{colortbl}
\usepackage{tcolorbox}

\definecolor{Gray}{gray}{0.93}

\newcommand{\ours}{LITA\xspace}
\newcommand{\ttoken}[1]{\texttt{<#1>}}
\newcommand{\cmark}{\ding{51}}%
\newcommand{\xmark}{\ding{55}}%

\def\paperID{****} % *** Enter the Paper ID here
\def\confName{CVPR}
\def\confYear{2024}

\title{LITA: Language Instructed Temporal-Localization Assistant}

\author{
De-An Huang,
Shijia Liao,
Subhashree Radhakrishnan,
Hongxu Yin,\\
Pavlo Molchanov,
Zhiding Yu,
Jan Kautz \\
NVIDIA \\
{\tt\small \{deahuang,shijial,subhashreer,dannyy,pmolchanov,zhidingy,jkautz\}@nvidia.com}
}

\begin{document}
\maketitle
\begin{abstract}
There has been tremendous progress in multimodal Large Language Models (LLMs). Recent works have extended these models to video input with promising instruction following capabilities. However, an important missing piece is temporal localization. These models cannot accurately answer the ``When?'' questions. We identify three key aspects that limit their temporal localization capabilities: (i) time representation, (ii) architecture, and (iii) data. We address these shortcomings by proposing Language Instructed Temporal-Localization Assistant (\ours) with the following features: (1) We introduce time tokens that encode timestamps relative to the video length to better represent time in videos. (2) We introduce SlowFast tokens in the architecture to capture temporal information at fine temporal resolution. (3) We emphasize temporal localization data for \ours. In addition to leveraging existing video datasets with timestamps, we propose a new task, Reasoning Temporal Localization (RTL), along with the dataset, ActivityNet-RTL, for learning and evaluating this task. Reasoning temporal localization requires both the reasoning and temporal localization of Video LLMs. \ours demonstrates strong performance on this challenging task, nearly doubling the temporal mean intersection-over-union (mIoU) of baselines. In addition, we show that our emphasis on temporal localization also substantially improves video-based text generation compared to existing Video LLMs, including a 36\% relative improvement of Temporal Understanding. Code is available at: \url{https://github.com/NVlabs/LITA}

\end{abstract}    
\section{Introduction}
\label{sec:intro}

Large language models (LLMs)~\cite{devlin2019bert,brown2020language,chowdhery2022palm,touvron2023llama,openai2023gpt4,touvron2023llama2}
have demonstrated impressive instruction following capabilities, and shown that language can be a universal interface for various tasks~\cite{openai2023gpt4,chowdhery2022palm}. These models can be further extended to multimodal LLMs  to process language and other modalities, such as image, video, and audio~\cite{alayrac2022flamingo,liu2023llava,zhu2023minigpt}. 

While most multimodal LLMs focus on images for visual content, several recent works introduce models that specialize in processing videos~\cite{damonlpsg2023videollama,luo2023valley,Maaz2023VideoChatGPT,2023videochat}. These Video LLMs preserve the instruction following capabilities of LLMs and allow users to ask various questions about a given video. However, one important missing piece in these Video LLMs is \emph{temporal localization}. When prompted with the ``When?'' questions, these models cannot accurately localize time periods, and often hallucinate irrelevant information~\cite{zhang2023hallucination}. Temporal localization is an important component that differentiates videos from images, and has been widely studied outside the context of instruction following LLMs~\cite{caba2015activitynet,jiang2014thumos,Damen2022RESCALING,sigurdsson2016hollywood}. It is thus crucial for Video LLMs to have temporal localization capabilities. 

\begin{figure}[t]
  \centering
  \includegraphics[width=1.0\linewidth]{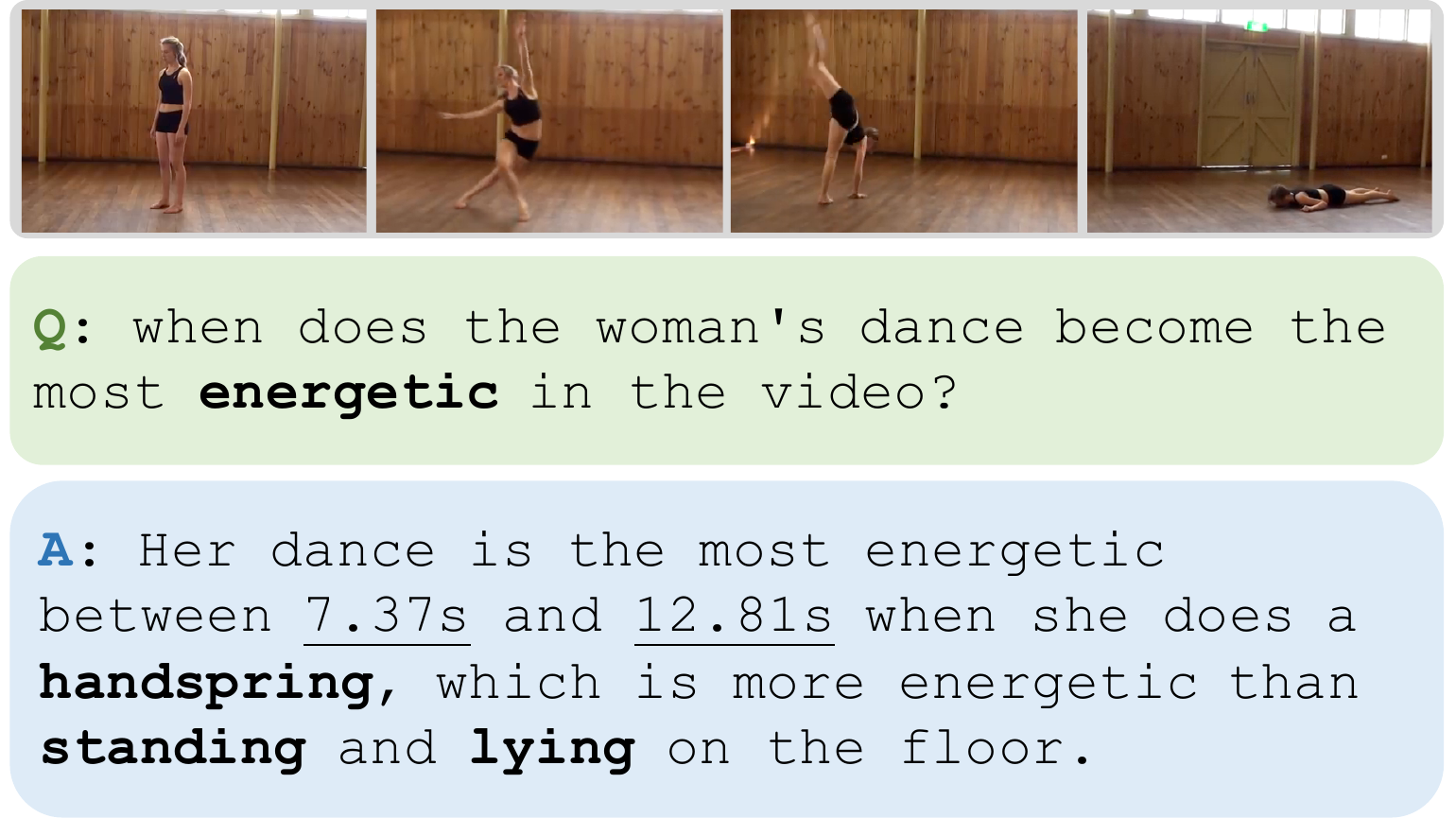}
  \caption{Example to illustrate our proposed Reasoning Temporal Localization (RTL). Instead of directly querying about an event, questions in RTL require further reasoning to answer. Here, the model needs to compare all activities in the video to find the timestamps of the most energetic activity (\ie, handspring).}
  \label{fig:fig1}
\end{figure}

We identify three key aspects that limit the temporal localization capabilities of existing Video LLMs: time representation, architecture, and data. First, existing models often represent timestamps as plain text (\eg 01:22 or 142sec). % to directly incorporate them into LLMs. 
However, given a set of frames, the correct timestamp still depends on the frame rate, which the model does not have access to. This makes learning temporal localization harder. 
Second, the architecture of existing Video LLMs might not have sufficient temporal resolution to interpolate time information accurately. For example, Video-LLaMA~\cite{damonlpsg2023videollama} only uniformly samples 8 frames from the entire video, which is insufficient for accurate temporal localization. 
Finally, temporal localization is largely ignored in the data used by existing Video LLMs. Data with timestamps are only a small subset of video instruction tuning data, and the accuracy of these timestamps is also not verified.

\textbf{Our Approach.} We address the aforementioned shortcomings of existing Video LLMs, and propose Language Instructed Temporal-Localization Assistant (\ours):
    (1) \emph{Time Representation}: We introduce \emph{time tokens} to represent relative timestamps and allow Video LLMs to better communicate about time than using plain text.
    (2) \emph{Architecture}: We introduce \emph{SlowFast tokens} to capture temporal information at fine temporal resolution to enable accurate temporal localization.
    (3) \emph{Data}: We emphasize temporal localization data for \ours. We propose a new task, Reasoning Temporal Localization (RTL), along with the dataset, ActivityNet-RTL, for learning this task.

The first important design of \ours is to use relative representation for time (\eg first 10\% of the video) instead of the absolute time representation with plain text (\eg 01:22). We divide a given video into $T$ equal length chunks, and introduce $T$ time tokens \ttoken{1} to \ttoken{T} to represent the relative time location in the video. During training and inference, these time tokens can be easily encoded and decoded from plain text timestamps given the length of the video. The start and end timestamps are well-defined by the time tokens given only the input video. This is in contrast to plain text timestamps. Without the frame rate, the correct absolute timestamp is ill-defined given just the video frames.

The second important design of \ours is to use densely sampled input frames from videos. It is unlikely to achieve accurate temporal localization with only sparsely sampled frames.
The challenge is that the LLM module inside Video LLMs cannot naively process large numbers of frames simultaneously due to context length limitation. Take LLaVA~\cite{liu2023llava} as an example. Each image is converted to 256 tokens, which are then fed into its LLM module as input. If we directly feed 100 frames to the LLM module, then that is $\text{256}\times\text{100}=\text{25600}$ tokens, which is already over the max context length for some LLMs~\cite{touvron2023llama2,vicuna2023}. Inspired by the SlowFast architecture for videos~\cite{feichtenhofer2019slowfast}, we instead consider two types of tokens, \textit{fast tokens} and \textit{slow tokens}, to address this efficiency issue. We generate \emph{fast tokens} at a high temporal resolution to provide the temporal information, while keeping the tokens per frame at a low number. On the other hand, we generate \emph{slow tokens} at a low temporal resolution, which allows us to use a higher number of tokens per frame to provide the spatial information.

Finally, We emphasize temporal localization data for \ours. We include dense video captioning~\cite{krishna2017dense} and event localization~\cite{yan2023unloc} in instruction tuning of \ours. These tasks include human annotated timestamps to promote accurate temporal localization. In addition to leveraging existing data and tasks, we further propose a new task, Reasoning Temporal Localization (RTL), along with the dataset, ActivityNet-RTL,  for training and evaluating this task. Answers to RTL questions can only be derived by utilizing world knowledge and temporal reasoning. \Cref{fig:fig1} shows an example. To answer the question: ``When does the woman's dance become the most energetic?'' the model needs to first recognize the woman's dance moves in the video, then reason about the most active part, and finally temporally localize the event (\ie handspring). In addition to the predicted timestamps, we further consider the \emph{explanation} provided by the model. Thus our new task not only assesses temporal understanding, but also requires strong reasoning capabilities that are unique to LLMs.

For the challenging RTL task, \ours doubles baseline's performance for temporal metrics (mIOU, Precision@0.5), while providing much better explanations. 
In addition to enabling accurate temporal localization, we show that our emphasis on temporal understanding also improves \ours's core Video LLM capabilities. \ours substantially improves all scores on a benchmark for video-based question answering~\cite{Maaz2023VideoChatGPT}. This includes a 22\% relative improvement for Correctness of Information, and a 36\% relative improvement for Temporal Understanding compared to existing Video LLMs.

\section{Related Work}
\label{sec:related}

\noindent\textbf{Multimodal Large Language Models.}
Large language models (LLMs)~\cite{openai2023gpt4,chowdhery2022palm} inspire recent works to address multimodal tasks by leveraging LLMs~\cite{wu2023next}. Some approaches add additional parameters inside LLMs, such as gated
cross-attention layers~\cite{alayrac2022flamingo,awadalla2023openflamingo,li2023mimicit} or adapter layers~\cite{zhang2023llama}, to adapt it to process multimodal inputs. Several works, on the other hand, only use modules, such as projection layers or Q-Formers, to project outputs of visual encoders to the input space of LLMs~\cite{liu2023llava,zhu2023minigpt,instructblip}. Recent works further expand multimodal LLM to visual grounding tasks, such as detection~\cite{kosmos-2,chen2023minigpt,liu2023improved} and segmentation~\cite{lai2023lisa}. The most related to ours is LISA~\cite{lai2023lisa}, which extends referring segmentation to reasoning segmentation. We share the same spirit and propose Reasoning Temporal Localization to jointly evaluate reasoning and temporal understanding.

\begin{figure*}[t]
  \centering
  \includegraphics[width=1.0\linewidth]{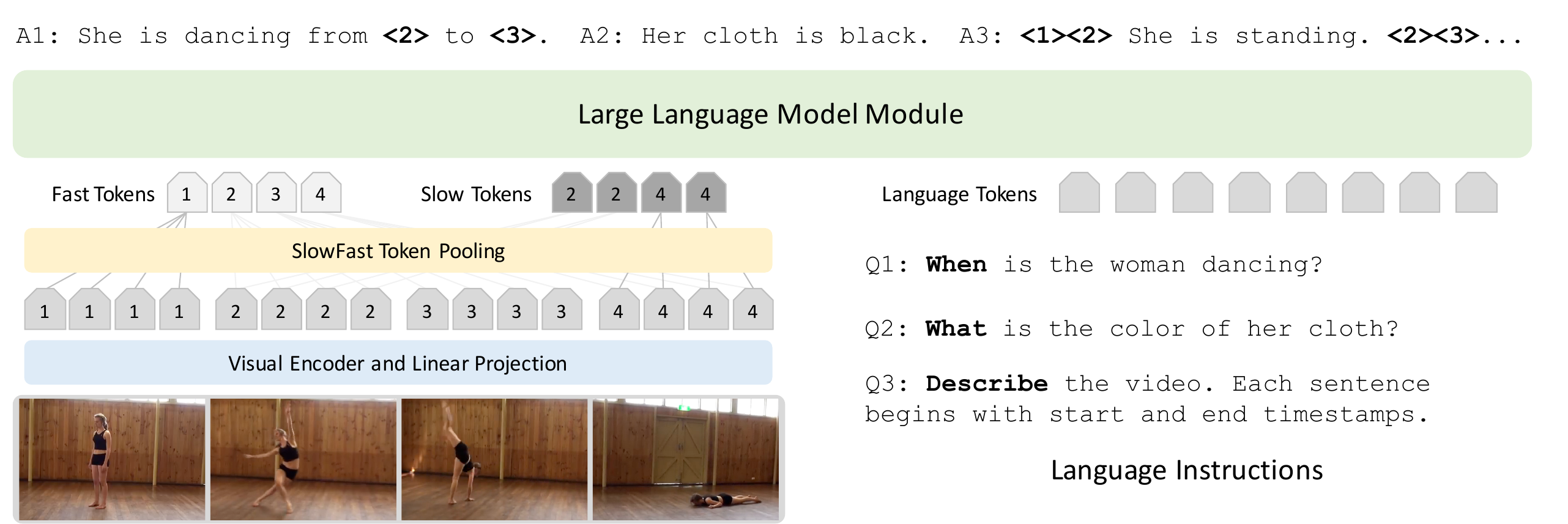}
  \caption{Overview of \ours. The input video frames are first encoded into visual tokens (numbered by frame), which are further processed by two pathways. The Fast Token pathway averages all the tokens in a frame to maintain a high temporal resolution. The Slow Token pathway sparsely samples frames to maintain a larger number of tokens per frame to provide spatial information. Timestamps are converted to time tokens \ttoken{1} to \ttoken{T}. This is important for better temporal localization learning. Various video tasks on the right can be converted to natural language visual question answering (\texttt{Q1-3} and \texttt{A1-3}) to jointly optimize \ours.}
  \label{fig:system}
\end{figure*}

\vspace{1mm}
\noindent\textbf{Video Large Language Models.} Building on the success of multimodal LLMs, several works extend image LLMs to Video LLMs~\cite{2023videochat,damonlpsg2023videollama,Maaz2023VideoChatGPT,luo2023valley}. These works mainly use the approach of projecting visual tokens to LLMs' input space using projection layers~\cite{Maaz2023VideoChatGPT,luo2023valley} or Q-Formers~\cite{damonlpsg2023videollama,2023videochat}. While these models show promise in descriptive questions and instructions, they still lack temporal localization capabilities. \ours is designed to address this shortcoming, while also improving downstream video tasks. Concurrent works~\cite{lin2023video,li2023llama,huang2023vtimellm,ren2023timechat,qian2024momentor} further improve existing VideoLLMs. The most related concurrent works to ours are VTimeLLM~\cite{huang2023vtimellm}, TimeChat~\cite{ren2023timechat}, and Momentor~\cite{qian2024momentor}. These works also aim to address temporal localization of Video LLMs. We further introduce the reasoning aspect to temporal localization.

\vspace{1mm}
\noindent\textbf{Temporal Localization in Videos.} The goal of temporal localization is to pinpoint activities within untrimmed video sequences on a temporal scale~\cite{yan2023unloc}. The target activities can be predefined action classes~\cite{jiang2014thumos,Damen2021PAMI} or events described by natural language~\cite{caba2015activitynet,ZhXuCoAAAI18}. Our goal of video temporal understanding is also related to various video tasks, such as dense video captioning~\cite{krishna2017dense,yang2023vid2seq,huang2020multimodal} and action segmentation~\cite{Kuehne12,tang2019coin,miech2019howto100m}. Models for these temporal tasks can have quite different design. Instruction following Video LLMs like \ours provide a way to unify these frameworks.

\section{Language Instructed Temporal-Localization}
\label{sec:method}

\ours enables temporal localization for Video LLMs by: (1) relative time representation with the \emph{time tokens}, (2) \emph{SlowFast tokens} to capture temporal information at fine temporal resolution, (3) multi-task training that includes accurate timestamps.  We will first introduce the overall architecture and discuss further details of individual components.

\subsection{Architecture}

An overview of \ours is shown in \Cref{fig:system}. We build on Image LLMs. In particular, we select LLaVA due to its simplicity and effectiveness~\cite{liu2023llava}. Note that \ours does not depend on the specific underlying Image LLM architecture and can be easily adapted to other base architectures.

Given a video, we first uniformly select $T$ frames and encode each frame into $M$ tokens. $T$ should be large enough to support the desired granularity of temporal localization. $T\times M$ is a large number that usually cannot be directly processed by the LLM module. Thus, we then use SlowFast pooling to reduce the $T\times M$ tokens to $T + M$ tokens.

The slow and fast tokens are projected by a linear layer and concatenated with the text tokens to use as input to the LLM module. The text tokens (prompt)  are processed to convert any referenced timestamps to specialized time tokens (\ttoken{1} to \ttoken{T}).
All the input tokens are then jointly processed by the LLM module sequentially.
We fine-tune the entire model with our reasoning temporal localization data (\Cref{sec:dataset}) along with other video tasks, such as dense video captioning and event localization. \ours learns to use time tokens instead of absolute timestamps. For temporal localization, we can then directly ask \ours the ``When'' questions (\eg ``When is she dancing?''). \ours would respond with time tokens (\eg ``She is dancing from \ttoken{2} to \ttoken{3}.''), which can then be converted to timestamps given the video length.

\subsection{Time Tokens}

We use a relative time representation instead of absolute timestamps in \ours. As shown in \Cref{fig:system}, the LLM module can only see the visual tokens (slow and fast) and the language tokens (text prompt). There is not enough information in this input space for the LLM module to infer the absolute timestamp because the frame rate is not known to the model in advance. A better way is to represent timestamps relative to the video length, thus removing the dependency on the frame rate. We divide the video into $T$ chunks and use $T$ specialized time tokens \ttoken{1} to \ttoken{T} for timestamps. Given a continuous timestamp $\tau$ and video length $L$, $\tau$ can be easily converted to time token \ttoken{t}, where $t=\texttt{round}(\tau(T-1)/L)+1$, and conversely \ttoken{t} can be converted back to $\tau = L(t-1)/(T-1)$. While this does introduce discretization error, it greatly simplifies the time representation with LLMs. Relative timestamp is also used in other temporally heavy video tasks, such as dense video captioning~\cite{yang2023vid2seq}. 

Given this time representation, many video tasks related to temporal localization can be transformed into language instructions and answers. For example, dense video captioning can be achieved by prompting the model with ``Describe the video. Each sentence begins with start and end timestamps.'' (\texttt{Q3} and \texttt{A3} in \cref{fig:system}). Standard event localization is also transformed to ``When does X happen?'' (\texttt{Q1} and \texttt{A1} in \cref{fig:system}). We also incorporate standard video question answering (\texttt{Q2} and \texttt{A2} in \cref{fig:system}). More details are discussed in \Cref{sec:training}.

\subsection{SlowFast Visual Tokens}

We have discussed how we discretize time in videos into $T$ steps in order to make Video LLMs better at reasoning about time. Still, the visual input should match the temporal resolution $T$ in order to achieve effective temporal processing. Ideally, one would need at least $T$ frames to temporally localize events with the resolution $T$. 
However, naively feeding all $T$ frames into the LLM module could be computationally prohibitive. In our experiment, we use $T=100$ and $M=256$ (CLIP ViT-L-14~\cite{radford2021learning}). This leads to 25600 tokens per video. 

Inspired by SlowFast models~\cite{feichtenhofer2019slowfast}, we consider two pathways to pool the $T\times M$ tokens for $T$ frames. The first is densely sampled \emph{fast tokens} to provide temporal information. We obtain $T$ fast tokens from $T$ frames by averaging all the tokens belonging to the same frame. The second is the sparsely sampled \emph{slow tokens} to maintain better spatial information. We select a spatial downsampling ratio of $s$, and uniformly select $s^2$ frames from the video. For each selected frame, we perform a $s\times s$ spatial average pooling to the $M$ tokens, which lead to $\frac{M}{s^2}$ slow tokens per frame. This leads to a total $M = \frac{M}{s^2} \times s^2$ slow tokens. We use $s=2$ in our experiments. This leads to a total of $T + M$ tokens to represent a video instead of $T \times M$ tokens.

\subsection{Training Tasks}
\label{sec:training}

In addition to architecture, training tasks and data also play an important role for LLMs. We emphasize temporal localization data and train \ours with the following five  tasks: (1) dense video captioning~\cite{krishna2017dense}, (2) event localization~\cite{yan2023unloc}, (3) video question answering~\cite{xiao2021next}, (4) natural language visual question answering~\cite{liu2023llava}, and (5) our proposed reasoning temporal localization. Temporal localization is a crucial component for three out of the five tasks (1, 2, and 5).

We now introduce each task in order. The first three tasks are standard video tasks and equip \ours with basic video understanding:

\vspace{1mm}
\noindent\textbf{Dense Video Captioning.} In dense video captioning~\cite{krishna2017dense}, each video is described by a set of sentences, and each sentence comes with the start and end timestamps of the event. Each sentence in dense video captioning can thus be represented as: 
\texttt{<start time> <end time> SENTENCE}. We then sort all sentences by its start time, and directly concatenate all sentences and timestamps. One example prompt to the model for this task is: ``Provide a detailed description of the given video. Each sentence should begin with the start and end timestamps.'' Other prompts are included in the supplementary materials. %\PM{what data are we using exactly here?}

\vspace{1mm}
\noindent\textbf{Event Localization.} In event localization, the goal is to temporally localize the event described by a sentence. We use a simple answer format: \texttt{<start time> <end time>}. One example prompt for this task is: ``When does ``\texttt{SENTENCE}'' happen in the video? Answer the question only using start and end timestamps.''

\vspace{1mm}
\noindent\textbf{Video Question Answering.} The question answering task is already represented as language instructions. However, answers in existing question answering datasets often consist of a single word or phrase because models for this task might not be able to generate longer text. We follow Liu~\etal~\cite{liu2023improved} and append the following prompt to the question: ``Answer the question using a single word or phrase.'' The goal is to provide the context for short answers so that it affects the model's text generation less.

\vspace{1mm}
\noindent\textbf{Natural Language Visual Question Answering.}
Training with the above three tasks gives \ours video understanding capabilities. However, we observe that models trained with only these tasks often provide short answers and lack natural language conversation capabilities. We thus also train \ours with natural language visual question answering or visual instruction tuning datasets~\cite{liu2023llava}. The goal is to improve the natural language conversation of \ours. We find that mixing instruction tuning datasets~\cite{liu2023llava} with standard video tasks improves \ours's conversation quality while maintaining good video understanding.

\vspace{1mm}
\noindent\textbf{Reasoning Temporal Localization.}
Finally, we also train \ours with our reasoning temporal localization task (details in \Cref{sec:dataset}). The answer to a reasoning temporal localization question consists of two parts: timestamps and explanation. We find it challenging for models to simultaneously output both of them without any example. Nevertheless, with some training data, \ours quickly pick up reasoning and temporal localization, and provide both the timestamps and explanation of its reasoning in answers.

\begin{figure*}[t]
  \centering
  \includegraphics[width=1.0\linewidth]{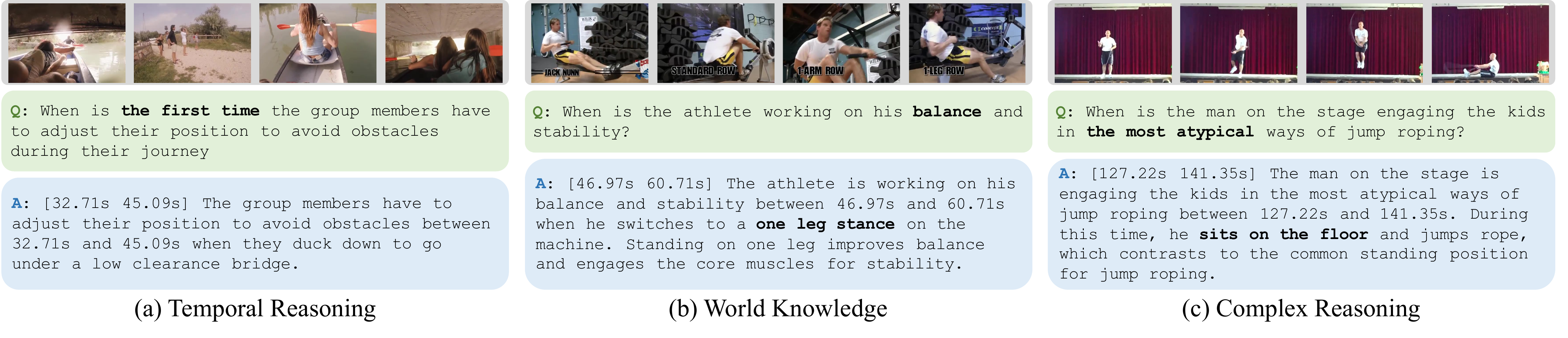}
  \caption{Examples from our ActivityNet-RTL dataset. RTL questions ask about events that are not explicitly described. The model needs to utilize reasoning or its world knowledge to answer. This is in contrast to standard temporal localization, which directly asks about the event of interest. For example in (c), standard temporal localization might directly ask: ``when does the man sit on the floor?''}
  \label{fig:dataset}
\end{figure*}

\section{Reasoning Temporal Localization}
\label{sec:dataset}

We now discuss further details of the proposed Reasoning Temporal Localization (RTL) task. Standard temporal localization does not fully leverage the potential of Video LLMs. One impressive aspect of LLMs is its reasoning capabilities. LLMs can even answer complex questions that involve multi-step reasoning. Therefore, we propose the RTL task to utilize both of Video LLMs' temporal understanding and reasoning capabilities.

\subsection{Problem Definition}

In \emph{reasoning} temporal localization, the query is still a ``when'' question that asks about the start and end timestamps of an event. The key difference compared to the standard temporal localization task is that the target event is not directly described in the question, and can only be inferred by reasoning and using world knowledge of the model. The answer to such a question thus consists of two parts: (1) the start and end timestamps of the target event, and (2) an explanation of the reasoning process the model goes through to derive the timestamps.

Some examples are shown in \Cref{fig:dataset} to better illustrate the idea. The answer format is: \texttt{[start end] Explanation}. In \Cref{fig:dataset}(a), the model not only has to localize ``adjust their position to avoid obstacles,'' but also needs to temporally reason about which instance happened earlier in the video. In \Cref{fig:dataset}(b), instead of directly asking about ``one-leg row,'' it asks about the workout targeting balance and stability. The model thus needs to utilize its knowledge of what kind of exercises are good for balance and stability. Finally, there are questions that require multi-step reasoning. In \Cref{fig:dataset}(c), the question asks about ``the most atypical ways of jump roping,'' which requires the model to understand what is typical and atypical for jump roping, and then temporally find the most atypical time period. A standard temporal localization task, in contrast, would just ask when the man is sitting on the floor. 

\subsection{ActivityNet-RTL Dataset}
\label{sec:anetrtl}

The above examples are from ActivityNet-RTL, a dataset curated by us for the Reasoning Temporal Localization (RTL) task. 
We build our dataset from the ActivityNet Captions dataset~\cite{krishna2017dense}, which annotates multiple events described by sentences in a video, and all the events are temporally localized with start and end timestamps. Consider the following toy example:
\vspace{-1mm}
\begin{small}
\begin{verbatim}
[00:00-00:10] A woman is standing.
[00:12-00:30] The woman is dancing.
[00:32-00:36] The woman is sleeping.
\end{verbatim}
\end{small}
\vspace{-1mm}
We then use this as context and ask GPT-4 to generate temporal localization questions that require further reasoning to answer. We also ask GPT-4 to simultaneously generate the answer that includes the queried start and end timestamps, along with the explanation about the reasoning process.

Using the above toy example, by seeing that the woman has done three activities, one possible reasoning temporal localization question is to ask ``When is she the least active?'' Since sleeping is the least active out of the three activities, the target time period is 00:32-00:36, the period when she is sleeping. This illustrates how GPT-4 can still generate interesting questions without seeing the actual video. We annotate few-shot examples for GPT-4 as in previous works to improve the generation quality~\cite{liu2023llava}. All of our prompts are included in the supplementary materials.

\vspace{1mm}
\noindent\textbf{Training Set Generation.} For our training set, we directly use the results generated by GPT-4 with 10,009 videos from the training split of ActivityNet-Captions. This leads to 33,557 Reasoning Temporal Localizaiton question-answer pairs. By inspecting the GPT generated results, we find that most of the questions are valid temporal localization questions given the context. The main shortcoming is that not all question-answer pairs require reasoning. Sometimes GPT-4 generates questions that directly ask about events that are already described by the dense video captions. However, we do hope that \ours can also answer these standard temporal localization questions correctly using natural language. We thus leave these questions in the training set.

\vspace{1mm}
\noindent\textbf{Evaluation Set Curation.} On the other hand, the evaluation set requires manual efforts otherwise we would end up with many non-reasoning questions. We start from the GPT-4 generated questions using a subset of the ActivityNet-Captions validation set, and manually remove non-reasoning questions. We also double check the timestamps and explanations in the answers. This leads to a total of 229 question-answer pairs for 160 videos.

\subsection{Metrics} 
\label{sec:metrics}

We consider three metrics for ActivityNet-RTL: mIOU, Precision@0.5, and GPT-4 Relative Scores. The first two metrics are for temporal localization, and the third metric is to evaluate the explanation capability. mIOU averages the intersection-over-union (IOU) between predicted and groundtruth start and end timestamps. Precision@0.5 measures the percentage of predictions that have over 0.5 IOU. We first average these two metrics per video, and then average over all videos in the evaluation set. This avoids over-weighting videos and time periods with more questions, as some time periods in a video have multiple questions. 

To evaluate the quality of the explanation, we follow the evaluation pipeline of LLaVA~\cite{liu2023llava} and leverage GPT-4 for evaluation. GPT-4 is asked to evaluate the helpfulness, relevance, accuracy, and level of details of the explanations, and give a score from 1 to 10. We ask GPT-4 to evaluate both the predicted and groundtruth explanations, and normalize the score for the prediction by the score of the groundtruth. For this metric, we directly average over all question-answer pairs as the explanations could be quite different even for questions about the same time period in the same video.

\section{Experiments}
\label{sec:experiments}
We evaluate \ours with both temporal localization and video tasks that do not involve temporal localization because most existing Video LLMs cannot handle temporal localization. In addition to our proposed Reasoning Temporal Localization, we further evaluate \ours on Video-based Text Generation Performance Benchmarking proposed by Maaz~\etal~\cite{Maaz2023VideoChatGPT}. This provides a holistic evaluation of \ours as a Video LLM and not just for temporal localization.

\subsection{Implementation Details}

\noindent\textbf{Architecture.} We uniformly sample 100 frames from a video, and use 100 time tokens \ttoken{1} to \ttoken{100} to represent timestamps. We use CLIP-L-14~\cite{radford2021learning} as the visual encoder, and Vicuna~\cite{vicuna2023} as the LLM module. We follow LLaVA's architecture and train a single linear layer for projection~\cite{liu2023llava}. We use 4 frames for slow tokens and use average pool window $s=2$. With 1 fast token per frame, this leads to a total of $100 + \frac{256}{4}\times 4 = 356$ tokens per video. 

\vspace{1mm}
\noindent\textbf{Training Datasets.} We discussed training tasks in \Cref{sec:training}. We now discuss the training datasets for each task. For dense video captioning and event localization, we use the training splits of ActivityNet-Captions~\cite{krishna2017dense} and YouCook2~\cite{ZhXuCoAAAI18}, which combine to around 11k videos. The event localization dataset can be generated from the dense video captioning dataset by using the caption as query and the timestamps as target. For video question answering, we use NExT-QA~\cite{xiao2021next} as it contains more complex questions. For image instruction tuning, we use LLaVA-150K~\cite{liu2023llava}. For reasoning temporal localization, we use the training split of our ActivityNet-RTL, which builts on the training split of ActivityNet-Captions.

\vspace{1mm}
\noindent\textbf{Training Setup.} For each of the five tasks, we randomly select 100K samples with replacement (total 500K). We then use batch size 128 and learning rate 2e-5 to train for 4k iterations. The training takes around 13 hours for 13B and 9 hours for 7B models using 8 A100 GPUs. The linear projection is initialized with the LLaVA pre-trained weights.

\subsection{Reasoning Temporal Localization Evaluation}
\label{sec:exp_model}

\begin{table}[t]
\caption{Results on ActivityNet-RTL. \ours substantially outperforms all baselines for all metrics. This shows the importance of our design choices. Interestingly the temporal localization accuracy also improves as we scale the model from 7B to 13B.
}
\label{tab:rtl_model}
\centering
\begin{tabular}{lcccc}
\toprule
Model            & Size & mIOU & P@0.5 & Score \\\midrule
LITA-7B             & 7B   & 24.1 & 21.2          & 44.0      \\\midrule
Video-LLaMA-v2~\cite{damonlpsg2023videollama}      & 13B  & --   & --            & 32.1      \\
Video-ChatGPT~\cite{Maaz2023VideoChatGPT}      & 13B  & --   & --            & 38.8      \\
Slow Tokens Only & 13B  & 14.6 & 11.8          & 32.2      \\
SlowFast Tokens  & 13B  & 17.5 & 14.5          & 34.1      \\
%\rowcolor{Gray}
LITA-13B             & 13B  & \textbf{28.6} & \textbf{25.9}          & \textbf{46.3}      \\\bottomrule
\end{tabular}
\end{table}

\begin{figure*}[t]
  \centering
  \includegraphics[width=1.0\linewidth]{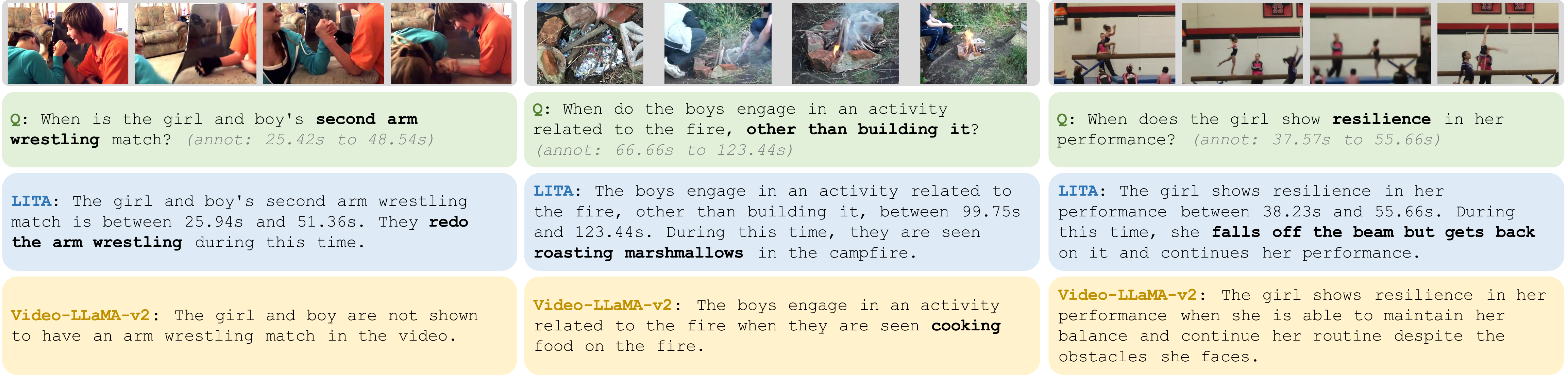}
  \caption{Qualitative results on ActivityNet-RTL. Overall, \ours not only more accurately localizes events in the video, but also provides sensible explanations with more details. In the first example, \ours correctly identifies the second arm wrestling. In the second example, \ours provides further details that they are roasting marshmallows. In the third example, \ours impressively recognizes that the girl ``falls off the beam but gets back'' and explains that this shows resilience. These are in contrast to the more generic answers by Video-LLaMA-v2.}
  \label{fig:qual_rtl}
\end{figure*}

\begin{table*}[t]
\caption{Video-based Text Generation Benchmarking results. \ours significantly outperforms existing Video LLMs including Video-LLaMA-v2~\cite{damonlpsg2023videollama} and Video-ChatGPT~\cite{Maaz2023VideoChatGPT} on all evaluated aspects. This shows that \ours not only enables accurate temporal localization, but also generally improves video understanding for Video LLMs.}
\label{tab:radar}
\centering
\begin{tabular}{lcccccc}
\toprule
              Model & {\scriptsize Correctness} & {\scriptsize Detail} & {\scriptsize Context} & {\scriptsize Temporal} & {\scriptsize Consistency}  & {\scriptsize Average}\\\midrule
LLaMA-Adapter & 2.03        & 2.32   & 2.30    & 1.98     & 2.15 & 2.16        \\
Video-LLaMA   & 1.96        & 2.18   & 2.16    & 1.82     & 1.79  &1.98    \\
Video-LLaMA-v2   & 2.36        & 2.42   & 2.74    & 1.83     & 2.12  &2.29      \\
VideoChat     & 2.23        & 2.50    & 2.53    & 1.94     & 2.24   &2.29     \\
Video-ChatGPT & 2.40        & 2.52   & 2.62    & 1.98     & 2.37  & 2.38      \\
LITA          & \textbf{2.94}        & \textbf{2.98}   & \textbf{3.43}    & \textbf{2.68}     & \textbf{3.19}  &\textbf{3.04} \\\bottomrule    
\end{tabular}
\end{table*}

We first evaluate on the newely proposed ActivityNet-RTL for reasoning temporal localization. Please refer to \Cref{sec:dataset} for dataset details and metrics. We use ``P@0.5'' for Precision@0.5 and ``Score'' for the GPT evaluation score for explanations. Other than variations of our model, we include Video-LLaMA-v2~\cite{damonlpsg2023videollama} and Video-ChatGPT~\cite{Maaz2023VideoChatGPT} for comparison. 
We observe that most of their outputs on ActivityNet-RTL omit any timestamps and thus mIOU and Precision@0.5 become absolute. Therefore, for these methods we only evaluate the GPT-Score.
In addition, we ablate the following variations for \ours, all of which are trained with the same five training tasks as \ours:

\vspace{1mm}
\noindent\textit{- ``Slow Tokens Only''} samples 4 frames from the video, and computes 64 tokens per frame. It does not use the fast tokens, and it also does not use time tokens for relative timestamps. This can be seen as naively implementing a Video LLM via the LLaVA architecture. 

\vspace{1mm}
\noindent\textit{- ``SlowFast Tokens''} additionally includes fast tokens (\ie 1 token per frame) compared to ``Slow Tokens Only''. This improves the architectural design for representing video, and should allow better temporal processing. 

\vspace{1mm}
\noindent\textit{- ``\ours''} is our full model that further includes time tokens to better represent timestamps compared to ``SlowFast Tokens.''  We consider two model sizes, 7B and 13B, for \ours to understand the effect of different model sizes.

\vspace{1mm}
\noindent\textbf{Importance of Our Model Components.} Results on ActivityNet-RTL are shown in \Cref{tab:rtl_model}. All metrics are averaged over three trials. \ours substantially outperforms all baselines for all metrics, and almost double mIOU and P@0.5 when compared to ``Slow Tokens Only'', which is considered as a naive extension of Image LLMs to Video LLMs. ``Score'' is assisted by GPT-4 to evaluate the quality of the explanation provided by the models. While we prompt GPT-4 to ignore the timestamps mentioned in the explanations,
we observe that the scores are still slightly affected by the timestamps. Nevertheless, \ours provides better explanations for its reasoning due to an overall better video understanding.

\vspace{1mm}
\noindent\textbf{\ours Gives Detailed and Accurate Explanations.} Qualitative results are shown in \Cref{fig:qual_rtl}. In the first example, \ours correctly localizes when the second arm wrestling happens. On the other hand, Video-LLaMA-v2 does not recognize that there is arm wrestling in the video. In the second example, while both models identify that the correct activity is cooking with fire, \ours provides much more accurate details, including the fact that they are roasting marshmallows, and the start and end time for that. Finally, in the third example, \ours impressively recognizes the event where the girl ``falls off the beam but gets back'' and correctly responds that this shows resilience in her performance. In contrast, Video-LLaMA-v2 gives a generic answer given the ``resilience'' prompt.

\vspace{1mm}
\noindent\textbf{Temporal Localization Scales with \ours Size.} One interesting observation is that the temporal localization quality of \ours also improves as we scale up the model from 7B to 13B (\Cref{tab:rtl_model}). One might think that scaling the model only improves language understanding, but our result shows that this could also improve  temporal reasoning and understanding for Video LLMs.

\begin{figure*}[t]
  \centering
  \includegraphics[width=1.0\linewidth]{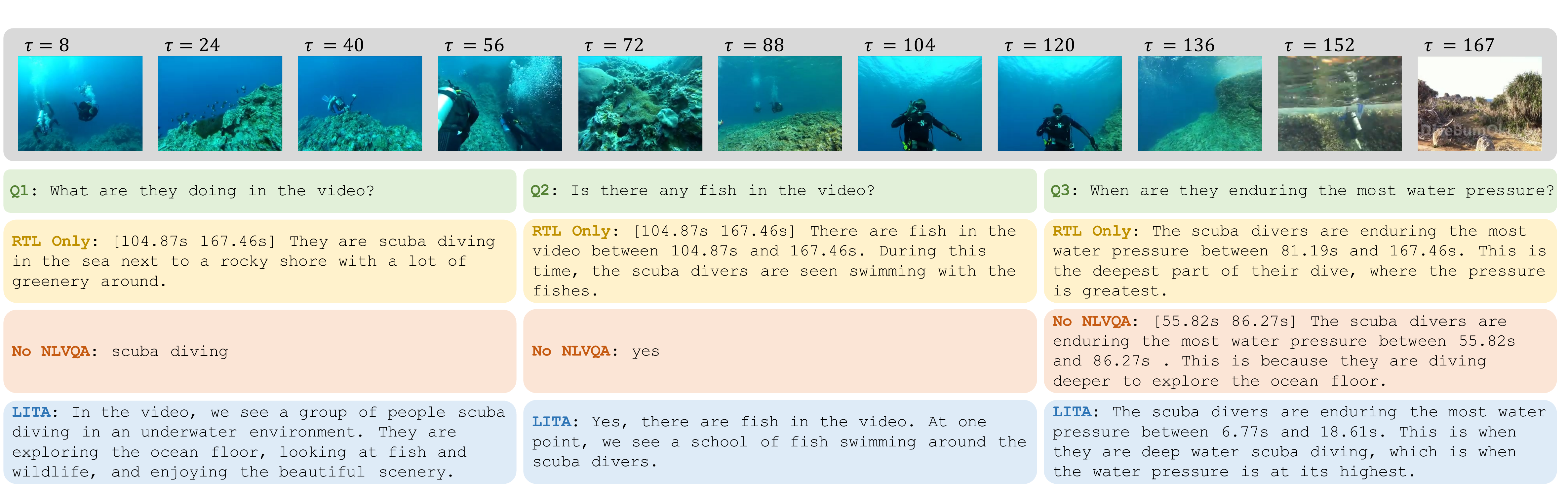}
  \caption{Qualitative results to analyze the effect of training tasks. $\tau$ is the continuous time in seconds. The questions are not from ActivityNet-RTL as we are also interested in questions other than reasoning temporal localization. ``RTL Only'' answers in the format for reasoning temporal localization, as this is the only training data it has. ``No NLVQA'' further includes standard video tasks and can correctly answer non-temporal questions (\texttt{Q1} and \texttt{Q2}). However, the answers are short due to the short answers in existing video question answering datasets. \ours improves both reasoning and natural language conversation by utilizing NLVQA datasets. 
  }
  \label{fig:qual_data}
\end{figure*}

\subsection{Video-Based Generation Evaluation}

In addition to our proposed reasoning temporal localization, we further evaluate \ours on standard evaluation for Video LLMs to better compare with existing Video LLMs. We use the ``Video-based Text Generation Performance Benchmarking'' proposed by Maaz~\etal~\cite{Maaz2023VideoChatGPT}. This benchmark selects videos from ActivityNet~\cite{caba2015activitynet} and annotates question-answer pairs with natural language. This is in contrast to existing video question answering benchmarks, where the answers are often limited to a single word or phrase. In this benchmark, there are specific questions like: ``What kind of tricks is the man performing while skating?'' or generic questions like ``Can you describe the video in detail?'' 
For evaluation, the benchmark uses GPT to measure the following aspects of Video LLMs' answers: Correctness of Information, Detail Orientation, Contextual Understanding, Temporal Understanding, and Consistency.

The results are shown in \Cref{tab:radar}. We compare \ours with LLaMA-Adapter~\cite{zhang2023llama}, Video-LLaMA~\cite{damonlpsg2023videollama}, VideoChat~\cite{2023videochat}, and Video-ChatGPT~\cite{Maaz2023VideoChatGPT}. 
Video-ChatGPT slightly outperforms other baselines including Video-LLaMA-v2. 
\ours significantly outperforms these two existing Video LLMs from all aspects. In particular, \ours achieves a 22\% improvement for Correctness of Information (2.94 vs.\ 2.40) and a 36\% relative improvement for Temporal Understanding (2.68 vs.\ 1.98). This shows that our emphasis of temporal understanding in training not only
enables accurate temporal localization, but also improves the video understanding of \ours. We hypothesize that temporal localization enables the model to learn more details about videos, leading to improved video understanding. A similar observation was made for Image LLMs~\cite{chen2023minigpt}, where joint training with grounding tasks also improved non-grounding text generation.

\subsection{Evaluating the Effects of Training Tasks}

We have analyze the effect of our model components in \Cref{sec:exp_model}, where we use all of our training tasks. Now we further analyze the effect of these training tasks for our model. We split the five tasks into three groups: RTL, Video, and NLVQA. ``RTL'' only includes the proposed reasoning temporal localization. Without training on our ActivityNet-RTL, the model would not output timestamps for us to evaluate temporal localization in many cases. The second group ``Video'' includes all the standard video tasks: dense video captioning, event localization, and video question answering. Using these video tasks, the model should learn better video understanding. Finally, ``NLVQA'' refers to the natural language visual question answering task to improve \ours's natural language conversation. We refer to training with just RTL as ``RTL Only,'' and training with both RTL and Video but without NLVQA as ``No NLVQA.'' Training with all three and thus all tasks is our proposed \ours.

\begin{table}[t]
\caption{Analysis of \ours's training tasks on ActivityNet-RTL. ``RTL'' is needed to predict both timestamps and explanations. ``Video'' includes standard video tasks to improve video understanding. ``NLVQA'' further improves reasoning and natural language generation capabilities of \ours.}
\label{tab:rtl_data}
\centering
\footnotesize
\begin{tabular}{l|ccc|ccc}
\toprule
Model & RTL & Video & NLVQA  & mIOU & P@0.5 & Score \\\midrule
RTL Only & \cmark & \xmark & \xmark & 26.6 & 20.9  & 43.5  \\
No NLVQA & \cmark & \cmark & \xmark & 26.9 & 23.5  & 44.9  \\
\ours & \cmark & \cmark & \cmark & \textbf{28.6} & \textbf{25.9}  & \textbf{46.3}  \\\bottomrule
\end{tabular}
\end{table}

\vspace{1mm}
\noindent\textbf{Results.} The results on ActivityNet-RTL are shown in \Cref{tab:rtl_data} and qualitative comparisons are shown in \Cref{fig:qual_data}. By only training on RTL, ``RTL Only'' does not have enough supervision to learn the task. This is reflected in both timestamp accuracy (P@0.5) and explanation quality (Score). In addition, for non-temporal questions (\texttt{Q1} and \texttt{Q2}) in \Cref{fig:qual_data}, the model cannot properly answer and always use the answer format for reasoning temporal localization. By adding standard video tasks in Video to training, ``No NLVQA'' improves all metrics compared to ``RTL Only''. Qualitatively in \Cref{fig:qual_data}, it can also answer \texttt{Q1} and \texttt{Q2} correctly. However, this capability mainly comes from the inclusion of video question answering datasets, which leads to short answers. \ours further improves by including NLVQA, which contains complex reasoning questions to improve its reasoning capabilities. More importantly, as shown in \Cref{fig:qual_data}, \ours now answers questions with natural language instead of short phrases.

\section{Conclusion}
\label{sec:conclusion}

We propose Language Instructed Temporal-Localization Assistant (\ours), which enables accurate temporal localization using Video LLMs. \ours demonstrates promising capabilities to answer complex temporal localization questions. At the same time, \ours substantially improves video-based text generation compared to existing Video LLMs even for non-temporal questions. 
This is the result of both our model design and data strategy. For model design, we propose time tokens to better represent the time and SlowFast tokens to efficiently process video inputs. Our experiments show the importance of these model components. For data strategy, we emphasize temporal localization data in training \ours. To achieve this, we propose the Reasoning Temporal Localization task and curate the ActivityNet-RTL dataset. Our results show that the inclusion of temporal localization data not only enables temporal localization for Video LLMs, but also improves the general video understanding capabilities.
We further analyze our training data and show the benefits of incorporating standard video tasks and image instruction tuning data to train Video LLMs.

{
    \small
    \bibliographystyle{ieeenat_fullname}
    \bibliography{main}
}

\end{document}